\documentclass[preprint,12pt]{elsarticle}

\usepackage{graphicx}%
\usepackage{multirow}%
\usepackage{amsmath,amssymb,amsfonts}%
\usepackage{amsthm}%
\usepackage{mathrsfs}%
\usepackage[title]{appendix}%
\usepackage{xcolor}%
\usepackage{textcomp}%
\usepackage{manyfoot}%
\usepackage{booktabs}%
\usepackage{algorithm}%
\usepackage{algorithmicx}%
\usepackage{algpseudocode}%
\usepackage{listings}%
\usepackage[english]{babel}
\usepackage{tabularx}
\usepackage{array}
\usepackage{ragged2e} 
\usepackage{enumitem}
\usepackage[hidelinks]{hyperref}
\usepackage{caption}

\journal{Journal of Biomedical Informatics}

\begin{document}

\begin{frontmatter}



\title{DispatchMAS: Fusing taxonomy and artificial intelligence agents for emergency medical services}


\fntext[eq]{These authors contributed equally to this work.}

\cortext[cor1]{Corresponding author.}

\author[sd]{Xiang Li\fnref{eq}}
\author[cuhk]{Huizi Yu\fnref{eq}}
\author[sd]{Wenkong Wang\fnref{eq}}

\author[psu]{Yiran Wu}
\author[stanford]{Jiayan Zhou}
\author[ucsb]{Wenyue Hua}
\author[cuhk]{Xinxin Lin}
\author[umac]{Wenjia Tan}
\author[nyu]{Lexuan Zhu}
\author[cuhk]{Bingyi Chen}
\author[broad,hku]{Guang Chen}
\author[stanford]{Ming-Li Chen}
\author[cams_pumc]{Yang Zhou}
\author[cams]{Zhao Li}
\author[stanford]{Themistocles L. Assimes}
\author[rutgers]{Yongfeng Zhang}
\author[psu]{Qingyun Wu}
\author[sd]{Xin Ma\corref{cor1}}\ead{maxin@sdu.edu.cn}
\author[usf]{Lingyao Li\corref{cor1}}\ead{lingyaol@sfu.edu}
\author[cuhk]{Lizhou Fan\corref{cor1}}\ead{leofan@cuhk.edu.hk}

\address[sd]{Shandong University, Jinan, Shandong, China}
\address[cuhk]{The Chinese University of Hong Kong, Sha Tin, NT, Hong Kong SAR, China}
\address[psu]{Pennsylvania State University, University Park, PA, United States}
\address[stanford]{Stanford University School of Medicine, Stanford, CA, United States}
\address[ucsb]{University of California, Santa Barbara, CA, United States}
\address[umac]{University of Macau, Macau SAR, China}
\address[nyu]{New York University, New York, NY, United States}
\address[broad]{Broad Institute of MIT and Harvard, Cambridge, MA, United States}
\address[hku]{The University of Hong Kong, Hong Kong SAR, China}
\address[cams_pumc]{Chinese Academy of Medical Sciences and Peking Union Medical College, Beijing, China}
\address[cams]{Chinese Academy of Medical Sciences, Beijing, China}
\address[rutgers]{Rutgers University, New Brunswick, NJ, United States}
\address[usf]{University of South Florida, Tampa, FL, United States}


\begin{abstract}

\textbf{Objective:} Emergency medical dispatch (EMD) is a high-stakes process challenged by caller distress, ambiguity, and cognitive load. Large Language Models (LLMs) and Multi-Agent Systems (MAS) offer opportunities to augment dispatchers. This study aimed to develop and evaluate a taxonomy-grounded, LLM-powered multi-agent system for simulating realistic EMD scenarios.

\textbf{Methods:} We constructed a clinical taxonomy (32 chief complaints, 6 caller identities from MIMIC-III) and a six-phase call protocol. Using this framework, we developed an AutoGen-based MAS with Caller and Dispatcher Agents. The system grounds interactions in a fact commons to ensure clinical plausibility and mitigate misinformation. We used a hybrid evaluation framework: four physicians assessed 100 simulated cases for ``Guidance Efficacy'' and ``Dispatch Effectiveness,'' supplemented by automated linguistic analysis (sentiment, readability, politeness).

\textbf{Results:} Human evaluation, with substantial inter-rater agreement (Gwet’s AC1 \(>\,\text{0.70}\)), confirmed the system’s high performance. It demonstrated excellent Dispatch Effectiveness (e.g., 94\% contacting the correct potential other agents) and Guidance Efficacy (advice provided in 91\% of cases), both rated highly by physicians. Algorithmic metrics corroborated these findings, indicating a predominantly neutral affective profile (73.7\% neutral sentiment; 90.4\% neutral emotion), high readability (Flesch 80.9), and a consistently polite style (60.0\% polite; 0\% impolite).

\textbf{Conclusion:} Our taxonomy-grounded MAS simulates diverse, clinically plausible dispatch scenarios with high fidelity. Findings support its use for dispatcher training, protocol evaluation, and as a foundation for real-time decision support. This work outlines a pathway for safely integrating advanced AI agents into emergency response workflows.
\end{abstract}


\end{frontmatter}



\section{Introduction}\label{sec1}

Emergency medical dispatch, where critical triage decisions are made based on limited caller information, is a pivotal step in prehospital care that directly impacts patient morbidity and mortality. National standards such as the Emergency Medical Dispatch Priority Reference System (EMDPRS)~\cite{maine2023_emdprs_bulletin, powerphone_emergency_call_handling_protocols} have been developed to promote consistency and safety across dispatch operations by guiding call-takers through structured, symptom-based protocols~\cite{slovis1985priority}. These systems aim to ensure timely and accurate identification of emergencies and the provision of life-saving pre-arrival instructions. 

However, emergency medical dispatch is inherently challenged by ambiguous symptom descriptions, linguistic diversity, high cognitive load, and caller distress—factors that can compromise decision accuracy and operational efficiency~\cite{gupta2024utilizing}. Despite the implementation of standardized guide cards and dispatcher training programs, prior studies have identified limitations in triage sensitivity for time-critical conditions, such as out-of-hospital cardiac arrest, and modest predictive value for advanced life support needs. Dispatchers and emergency medical services (EMS) leadership increasingly view these limitations as areas where intelligent decision support could augment human judgment, particularly during complex or uncertain calls~\cite{otal2024llm}.

To address these gaps, we present a fully autonomous, multi-agent LLM system that simulates both caller and dispatcher roles. The agents are constrained at every turn by a clinically curated dispatch taxonomy and a shared fact commons (a validated knowledge base of clinical/operational rules), enabling procedurally aligned, clinically faithful, and linguistically diverse, context-controlled conversations. Unlike approaches that mine static narratives or rely on human-in-the-loop role-play ~\cite{rahman2023ems, chen2025sim911}, our system generates dynamic scenarios for end-to-end evaluation and protocol stress-testing.

We assess performance with a hybrid evaluation strategy combining expert clinician assessments with automated metrics of linguistic quality, including sentiment, readability, and politeness. This dual framework enables evaluation of both clinical plausibility and communication fidelity. The system is tested across a range of standardized emergency scenarios to evaluate realism, alignment with clinical intent, and the communicative appropriateness of AI-generated responses. Therefore, the primary contribution of this study is threefold: 
(1) we construct a comprehensive EMS taxonomy and fact commons that ground dispatcher–caller simulations; 
(2) we develop an LLM-based multi-agent dispatch system aligned with these structures; and 
(3) we design a hybrid human–algorithm evaluation framework. 
Key findings supporting these contributions include high operational quality (e.g., 94\% correct external-agent contact, 97\% call-back instruction, 91\% advice provided) and strong communication metrics (73.7\% neutral sentiment , 90.4\% neutral emotions, Flesch Reading Ease 80.9, and 60.0\% polite with 0\% impolite).

{
\centering
\renewcommand{\arraystretch}{1.2}

\begin{tabularx}{\linewidth}{>{\bfseries}p{0.28\linewidth} X}
\toprule
\multicolumn{2}{l}{\textbf{Statement of significance}} \\ 
\midrule
\textbf{Summary} & \textbf{Description} \\
\midrule
Problem or issue &
Emergency medical dispatch (EMD) is a high-stakes process where dispatcher performance is limited by cognitive load, caller distress, and ambiguous information. Scalable, clinically grounded tools for dispatcher training, protocol evaluation, and decision support are lacking.\\
What is already known &
LLMs and Multi-Agent Systems (MAS) have been explored for EMS tasks such as call classification or simulations (e.g., Sim911). However, these systems are not fully autonomous and lack grounding in clinical taxonomies, limiting their ability to generate diverse, clinically faithful scenarios.\\
What this paper adds &
This study introduces DispatchMAS, a fully autonomous multi-agent system simulating EMD interactions by grounding LLM-based agents in a clinical taxonomy and fact commons. We present a framework for generating diverse, clinically plausible scenarios and a hybrid evaluation methodology.\\
Who would benefit from the knowledge in this paper &
EMS leaders, training programs, and AI researchers benefit from this methodology for simulations, improving training and protocol design.\\
\bottomrule
\end{tabularx}
}

\section{Related Work}
\subsection{Large Language Models in Emergency Medical Services}
The emergence of Large Language Models (LLMs) ~\cite{yu2024review} has transformed the landscape of clinical natural language processing. In emergency contexts, LLMs have been used to classify call urgency~\cite{otal2024llm}, assist with triage decisions~\cite{arslan2025evaluating}, and generate structured summaries from unstructured dispatcher-caller dialogues. Evaluation studies have shown moderate-to-strong agreement between LLM triage outputs and human paramedic decisions, suggesting early feasibility of integrating LLMs into prehospital workflows for clinical decision support~\cite{shekhar2025use}. Other work has demonstrated high classification accuracy of emergency versus non-emergency calls using prompt-engineered LLMs trained on call transcripts and medical scenarios ~\cite{akaybicen2024machine}. More recently, multimodal applications of LLMs have emerged in the form of wearable cognitive assistants that leverage speech, vision, and biometric inputs to assist emergency responders during field operations~\cite{10562157}.

\subsection{Multi-Agent Systems in Prehospital Care}
Multi-Agent Systems (MAS) have also been explored as a complementary framework to address the coordination and operational complexity of EMS. MAS architectures enable multiple autonomous agents, representing stakeholders such as dispatch centers, ambulance teams, hospitals, or public safety systems, to collaborate in real time. These agents communicate through defined protocols to manage triage workflows, resource allocation, and route optimization, improving the efficiency and scalability of EMS systems. Earlier implementations of MAS in pre-hospital care have demonstrated benefits in task distribution and situational awareness \cite{safdari2017multi, iliashenko2021implementing}. More recent work has integrated MAS with reinforcement learning to dynamically adapt responder deployments based on spatial demand patterns, enhancing response coverage and reducing travel times in densely populated urban areas \cite{pmlr-v235-sivagnanam24a,fu2025modelling}.

\subsection{Fusion of LLMs and MAS in Simulation and Decision Support}
The convergence of MAS and LLMs introduces an opportunity to synthesize structured coordination and natural language intelligence~\cite{altermatt2025multiagent}. Similar explorations of LLMs in other high-stakes domains, such as national security applications~\cite{caballero2025large}, further highlight the importance of safety, accountability, and reliability in mission-critical contexts. In simulation-based EMS training, LLM-powered agents have been deployed to generate realistic, multilingual caller personas~\cite{yu2024aipatient} and to simulate dispatcher interactions, enhancing both communication fidelity and scenario variability~\cite{chen2025sim911, hartman2024developing}. Similarly, agent-based LLM systems have been proposed to support emergency department workflows through role-specific reasoning agents, such as triage nurses, physicians, and documentation assistants, collaboratively processing patient data in real time~\cite{HanChoi2025_AAIML_KTAS}. These architectures exemplify the potential for multi-agent LLM systems to augment human operators without replacing clinical judgment, particularly when embedded within human-in-the-loop frameworks.

Nevertheless, operational challenges persist. LLMs can hallucinate factual content—especially in long-context settings—and their outputs may be sensitive to prompt phrasing and domain drift~\cite{kim2025hallucinations}. Current efforts address these issues via instruction fine-tuning on curated EMS datasets and retrieval-augmented generation that anchors outputs to structured dispatch taxonomies~\cite{akaybicen2024machine,10562157,chen2025sim911,HanChoi2025_AAIML_KTAS}. In parallel, MAS deployments must integrate with existing EMS infrastructure, respect jurisdiction-specific triage protocols, and provide explainable agent behaviors for incident review.

Against this backdrop, prior systems have taken two main paths. Some concentrate on mining static EMS narratives with domain-specific models (e.g., entity and relation extraction) rather than producing interactive scenarios~\cite{rahman2023ems}. Others move toward simulation but retain a human-in-the-loop design, using LLM callers to train dispatchers with RAG-supported, context-controlled workflows~\cite{chen2025sim911}.

Our approach advances beyond these directions by instantiating both caller and dispatcher as LLM agents, each constrained by a dispatch taxonomy and fact commons. This design yields autonomous, procedurally aligned, and scalable conversations that extend beyond human role-play, supporting protocol evaluation today and laying foundations for decision support.


\section{Methods}\label{sec2}
\begin{figure}[htbp]
  \centering
  \includegraphics[width=\linewidth]{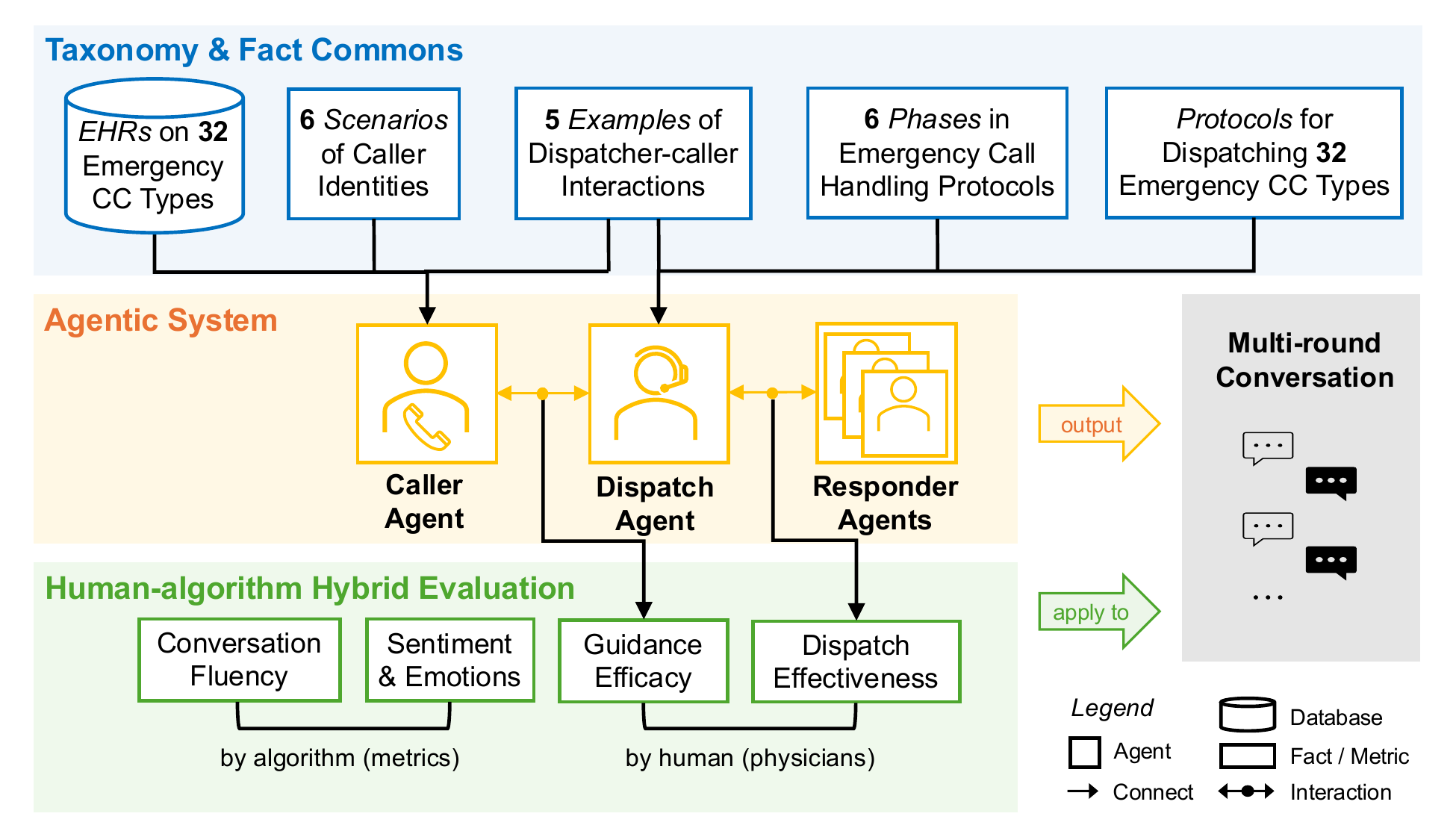} 
  \caption{Overview of the three-phase methodology: (a) Taxonomy \& Fact Commons, (b) Agentic System, and (c) Human–algorithm Hybrid Evaluation.}
  \label{fig:overview}
\end{figure}

This study employs a systematic three-phase methodology designed to ensure clinical grounding, technical robustness, and rigorous evaluation of the proposed dispatch simulation system. As illustrated in Figure~\ref{fig:overview}, the first phase focuses on constructing a comprehensive taxonomy and fact commons that formally define 32 chief complaints (CCs), six caller identities, and a six-phase call-handling protocols. The second phase translates this structured knowledge into an LLM-based MAS, enabling dynamic interactions between caller, dispatcher, and responder agents. The third phase establishes a hybrid evaluation framework, combining physician adjudication of guidance efficacy and dispatch effectiveness with algorithmic assessments of conversation fluency, sentiment and emotions, readability, and politeness. Together, these phases provide an integrated pathway for developing and validating AI agents in emergency medical services.

\subsection{A taxonomy and fact commons for emergency medicine }\label{subsec1}
We developed a comprehensive taxonomy and fact commons to systematically organize the diverse scenarios encountered in emergency medical dispatch. This structured knowledge base underpinned our simulation system by ensuring that language model agents operated within clinically grounded and procedurally accurate representations of emergency calls~\cite{CUI2025104861}. To construct the taxonomy, we identified 32 distinct CCs derived from the Emergency Medical Dispatch: National Standard Curriculum published by the National Highway Traffic Safety Administration (NHTSA)\cite{national1997emergency, nhtsa1996emtbasic} and the Health Resources and Services Administration (HRSA), with additional guidance from the Maternal and Child Health Bureau. These CCs spanned a wide range of emergency types, including medical, traumatic, environmental, and obstetric events, and were selected for their representativeness and relevance to dispatcher workflows\cite{nhtsa1996_emd_nsc}.

For each CC, we curated structured information to support both simulation and interpretation. The detailed field schema, curation principles, and full categorization of this taxonomy are available in Supplementary Supplementary Method 1. This included the clinical background and typical etiologies, the most frequently reported symptoms and situational patterns, standard pre-arrival instructions, special considerations for pediatric and maternal cases, and integration of relevant local and national dispatch protocols (Supplementary Table 6)~\cite{lin2023multimodal}. Together, these elements ensure clinical fidelity and provide a comprehensive fact commons that enables agents to respond in ways consistent with best practices in emergency medicine.

To simulate realistic emergency interactions, we further designed six prototypical caller identities reflecting the diversity of real-life dispatch scenarios~\cite{ASTM_F1258_2000}:
\begin{enumerate}[leftmargin=*, labelindent=0pt]
    \item \textbf{Patient}: the individual, who may be acutely symptomatic and have limited capacity to communicate;
	\item \textbf{Bystander}: a third-party bystander, present but unfamiliar with the patient’s background;
	\item \textbf{Family/Associate}: a family member or close associate, often introducing emotional urgency;
	\item \textbf{Multiple Callers}: multiple concurrent callers, potentially providing conflicting information;
	\item \textbf{Involved Party}: an individual possibly responsible for the emergency; and
	\item \textbf{Limited-Proficiency Caller}: a linguistically mismatched caller, whose primary language differs from that of the dispatcher.
\end{enumerate}
These identities were grounded in clinical narratives drawn from the MIMIC-III electronic health record dataset~\cite{li2024scopingreviewusinglarge}, enabling the system to reflect realistic expressions of symptoms, caller behaviors, and information quality.

In parallel, we encoded the standard progression of an emergency call using a six-phase protocol widely adopted in dispatcher training and operations~\cite{nhtsa1996_emd_nsc}:
\begin{enumerate}
    \item \textbf{Initial Intake}: where call handlers collect essential information such as location, patient identity, and nature of the emergency;
    \item \textbf{Scene Condition Assessment}: eliciting contextual details to evaluate environmental safety and condition severity;
    \item \textbf{Dispatch}: initiating appropriate response units based on collected information;
    \item \textbf{Provision of Real-Time Updates}: as new details emerge;
    \item \textbf{Delivery of Pre-arrival Instructions}: such as CPR and bleeding control to stabilize the patient prior to responder arrival; and
    \item \textbf{Call Closure}: formally concluding the interaction once professional responders assume care.
\end{enumerate}


\subsection{An LLM-based agentic EMS dispatch system }\label{subsec2}

\begin{figure}[htbp]
  \centering
  \includegraphics[width=\linewidth]{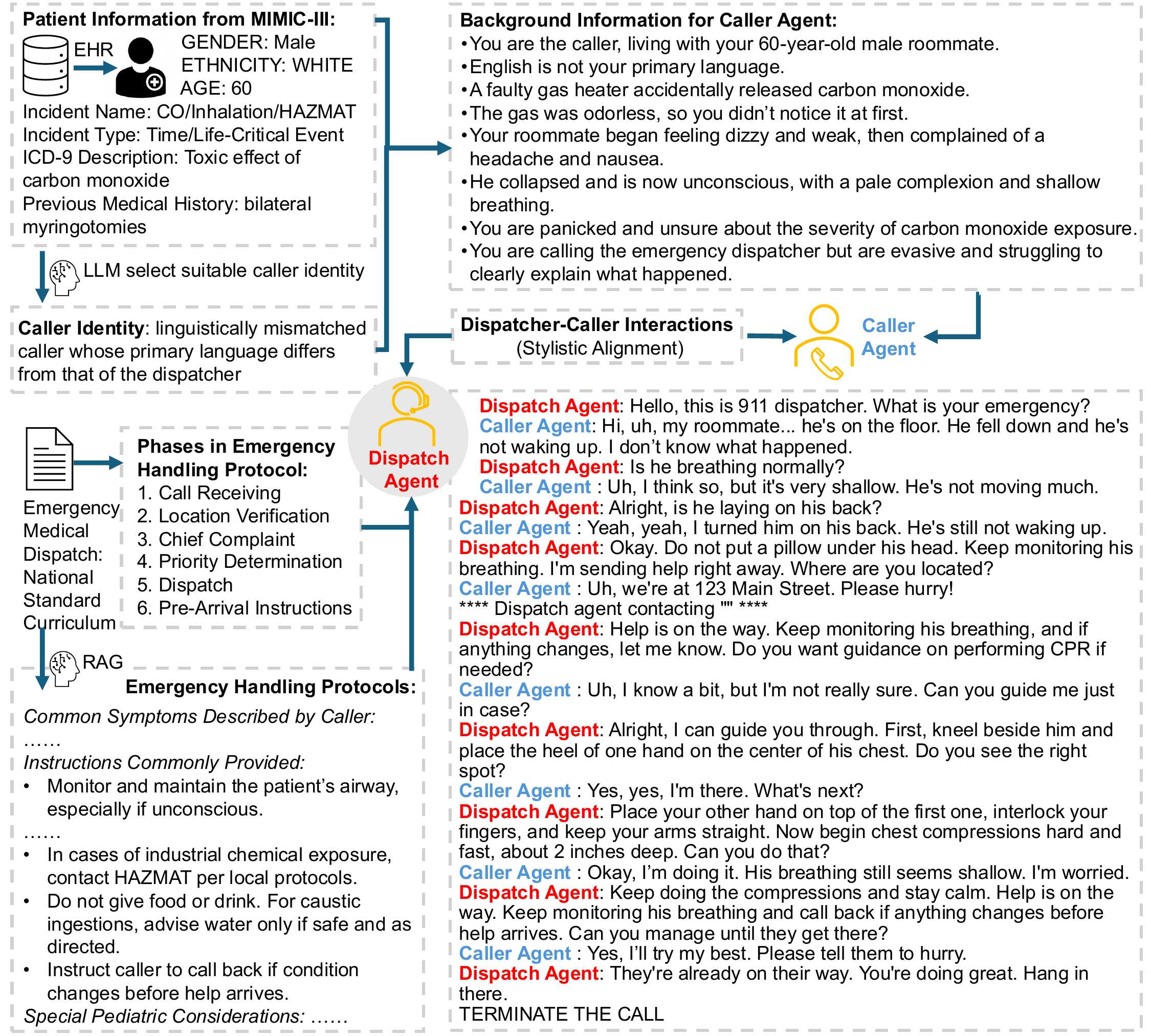}
  \caption{Schematic overview of the LLM-based multi-agent emergency medical dispatch (EMD) simulation system. Patient information derived from MIMIC-III electronic health records (EHR) seeds realistic caller backgrounds and identities. Knowledge from the Emergency Medical Dispatch National Standard Curriculum—including chief complaint taxonomy, handling protocols, and call-handling phases—grounds both caller and dispatcher agents. The pipeline illustrates caller scenario generation and identity selection, dispatcher–caller interactions with turn-level chief-complaint classification and protocol retrieval, optional escalation to auxiliary agents, and delivery of pre-arrival instructions, as exemplified by the representative call transcript.}
  \label{fig:emd_system}
\end{figure}

Building on the structured taxonomy and fact commons described in Section~\ref{subsec1}, our second major contribution is the development of a multi-agent emergency medical dispatch (EMD) simulation system powered by LLMs. As illustrated in Figure~\ref{fig:emd_system}, the system integrates structured medical knowledge, realistic caller identities, and standardized dispatch protocols to simulate end-to-end emergency call scenarios. Two core agents, a caller agent and a dispatcher agent, interact dynamically within the AutoGen framework~\cite{wu2024autogen} to reproduce the real-world 911 calls. The specific component versions, including the core LLMs and software frameworks, are documented in Supplementary Table 1 to ensure reproducibility.

Building on the structured taxonomy and fact commons, our second major contribution is the development of a multi-agent EMD simulation system powered by LLMs. The system simulates emergency call scenarios through dynamic interactions between two core agents, a caller agent and a dispatcher agent, implemented using the AutoGen framework. 

A well-known limitation of LLMs is their tendency to generate plausible but inaccurate or fabricated responses, commonly referred to as hallucinations\cite{huang2025survey}. To mitigate this risk, both agents are grounded in the fact commons introduced in Section~\ref{subsec1}. By constraining the generative process with structured, clinically relevant knowledge at each conversational turn, the system minimizes misinformation and enhances clinical realism.

The caller agent is designed to simulate diverse and naturalistic emergency calls. For each case, we generate a narrative background scenario rather than directly exposing structured patient data. Using profiles derived from MIMIC-III~\cite{johnson2016mimic}, the LLM selects the most contextually appropriate caller identity from six predefined roles—patient, bystander, family member or associate, multiple callers, potentially responsible individual, or linguistically mismatched caller—while excluding logically inconsistent options (e.g., an unconscious patient acting as the caller). The generated scenario includes contextual details such as caller–patient relationship, setting (e.g., home or public space), time of day, and observable symptoms (see Supplementary Table 7 for sampling variables). This approach achieves two objectives: it ensures realistic alignment of caller identity with clinical context and avoids overly precise or implausibly clinical descriptions that would not occur in real conversations. The construction workflow, sampling variables, and critical safety-first consistency rules governing the Caller Agent generation are detailed in Supplementary Supplementary Method 2. To further approximate authentic call dynamics, the caller agent utterances are calibrated with linguistic samples from five real 911 transcripts, allowing the system to reproduce the ambiguity, urgency, and emotional tone typical of emergency calls.

The dispatcher agent is initialized with standardized Emergency Call Handling Protocols and the same authentic call transcripts used to calibrate the caller agent. At each conversational turn, a lightweight classifier reviews the dialogue history and assigns one of the 32 CCs categories defined in the taxonomy, or a fallback category named ‘Lack of Information’ if insufficient evidence is available. Once a CC is determined, we employ a RAG mechanism. This process dynamically retrieves relevant procedural and clinical instructions specific to the identified CC from our knowledge base and integrates them into the dispatcher agent's prompt. This ensures structured questioning, appropriate red-flag coverage, and accurate pre-arrival guidance. The system's turn-level RAG loop and the full prompt library, including global hard constraints designed to ensure safe and coherent interactions, are described in Supplementary Supplementary Method 3 and Supplementary Supplementary Method 4. The full prompts governing the dispatcher agent, caller agent, CC classifier, and RAG injection template are provided in Supplementary Tables 2–5.

In addition, the system supports multi-agent escalation and external tool integration. Each CC protocol is annotated with potential auxiliary resources, such as the EMDPRS or external agencies (e.g., police, fire services). Upon CC confirmation, the dispatcher can invoke these resources through function calls. In this study, responses from auxiliary agents were mocked by LLMs to maintain dialogue flow, but the modular design allows plug-and-play integration with real APIs or agent logic in future deployments.

Taken together, the agentic system operationalizes the end-to-end workflow depicted in Figure~\ref{fig:emd_system}: structured case metadata and EHR-derived narratives seed realistic caller scenarios; the dispatcher iteratively classifies CCs and retrieves protocol-bound decision steps; pre-arrival instructions are delivered when criteria are met; and, when applicable, function calls trigger auxiliary agents (e.g., EMDPRS, police, fire). By unifying grounded medical knowledge, narrative caller modeling, dynamic CC-conditioned prompting, and modular escalation, the framework yields high-fidelity simulations that mirror real 9-1-1 interactions while maintaining clinical safety. This platform therefore supports controlled training and evaluation today, and provides a plug-and-play pathway to future operational integration with live EMS systems.

\subsection{Human–Algorithm Hybrid Evaluation Framework}\label{subsec3}

We evaluated the AI-enhanced dispatch system using a hybrid framework that combined structured human expert assessment with algorithmic communication–quality metrics. This dual-track design addressed two complementary goals: \emph{clinical validity}, which referred to whether the system adhered to emergency medical dispatch protocols and provided safe, appropriate guidance; and \emph{communicative robustness}, which referred to whether the system maintained clarity, calmness, and professionalism under stressful conditions.

\subsubsection*{Integrated Assessment}
Human and algorithmic evaluations provide complementary perspectives. Human evaluation ensures clinical appropriateness, adherence to emergency dispatch protocols, and practical utility, while algorithmic evaluation enforces clarity, neutrality, and professionalism at scale. In combination, these methods establish a robust framework for validating both the clinical and communicative performance of the AI dispatch system, thereby supporting safe deployment and iterative refinement.

\subsubsection*{Human Evaluation}
Two domains were defined for human-centered evaluation: \textbf{Guidance Efficacy} and \textbf{Dispatch Effectiveness}. Guidance Efficacy captures whether advice is given, and whether its amount and helpfulness are sufficient for callers. Dispatch Effectiveness focuses on operational quality, including the appropriateness and relevance of questions, contacting auxiliary services, and delivery of safety reminders. The structured questionnaire is shown in Table~\ref{tab:evaluation_questionnaire_ver2}.

\newcolumntype{L}[1]{>{\RaggedRight\arraybackslash\hspace{0pt}}p{#1}}
\newcolumntype{C}[1]{>{\centering\arraybackslash}p{#1}}
\newcolumntype{Y}{>{\RaggedRight\arraybackslash\hspace{0pt}}X}

\begin{table}[htbp]
\centering
\caption{Overview of Evaluation Questionnaire}
\label{tab:evaluation_questionnaire_ver2}

\small
\setlength{\tabcolsep}{6pt}
\renewcommand{\arraystretch}{1.2}

\begin{tabularx}{\textwidth}{@{} L{3.7cm} Y C{2.8cm} @{}}
\toprule
\textbf{Category} & \textbf{Question} & \textbf{Answer Type} \\
\midrule

\multicolumn{3}{@{}l}{\textbf{Guidance Efficacy}}\\
\addlinespace[2pt]

Advice given &
Did the Dispatcher provide advice to the Caller? &
Binary (Yes/No) \\

Satisfaction with amount of advice &
Was the amount of advice provided by the Dispatcher adequate? &
Ordinal (1--5)\textsuperscript{1} \\

Helpfulness of advice, if given &
Was the advice given by the Dispatcher helpful in assisting the Caller during the emergency? &
Ordinal (1--5)\textsuperscript{1} \\
\addlinespace[2pt]

\multicolumn{3}{@{}l}{\textbf{Dispatch Effectiveness}}\\
\addlinespace[2pt]

Number of questions asked and answered &
Was the number of questions asked and answered between the Dispatcher and Caller reasonable? &
Ordinal (1--5)\textsuperscript{1} \\

Relevance of questions asked and answered &
Did the Dispatcher ask relevant questions to identify the medical issue? &
Ordinal (1--5)\textsuperscript{1} \\

Contact the correct potential other agents &
Did the Dispatcher successfully contact the correct potential other agents? &
Binary (Yes/No) \\

Told to call back if necessary &
Did the Dispatcher advise the Caller to call back if necessary? &
Binary (Yes/No) \\
\bottomrule
\end{tabularx}

\vspace{3pt}
\begin{minipage}{\textwidth}
\footnotesize\emph{Note}. \textsuperscript{1} Anchors for all ordinal items: 
1 = strongly dissatisfied, 2 = dissatisfied, 3 = acceptable, 4 = satisfied, 5 = very satisfied.
\end{minipage}
\end{table}

We generated 100 simulated emergency call scenarios covering diverse CCs and caller profiles. Four licensed physicians who completed emergency medicine training/rotations and have experience with acute triage served as expert raters. Each rater independently reviewed an assigned subset of cases using the questionnaire, while 20 cases were annotated by all four physicians to assess inter-rater reliability. Raters were blinded to model version and trained on a calibration set to standardize interpretation. The evaluation interface, as shown in Supplementary Figure. 3, consist of two panels: the left panel displays the dialogue between the caller agent and the dispatch agent, while the right panel contains the standardized evaluation options. Physicians submit their assessments through this interface.


\subsubsection*{Algorithmic Evaluation}
To complement expert judgment, we assessed four communication-related dimensions: \textbf{Sentiment Analysis} (balanced and reassuring tone), \textbf{Emotion Classification} (appropriate emotional calibration), \textbf{Readability Assessment} (clarity under stress), and \textbf{Politeness Evaluation} (professionalism). Categories, taxonomies, and models are summarized in Table~\ref{tab:algorithmic_metrics_ver2}. The specific model checkpoints and specifications used for this analysis are detailed in Supplementary Table 8.

\begin{table}[htbp]
\centering
\caption{Algorithmic Evaluation Metrics and Models}
\label{tab:algorithmic_metrics_ver2}
\begin{tabularx}{\textwidth}{@{} X l X l @{}}
\toprule
\textbf{Category} & \textbf{Typology} & \textbf{Taxonomy} & \textbf{Model / Calculation} \\
\midrule
Sentiment Analysis & Multi-Class & Positive, negative, and neutral & RoBERTa-base \\
\addlinespace
Emotion Classification & Multi-Class & Disgust, joy, sadness, anger, fear, surprise, and neutral & E-DistilRoBERTa \\
\addlinespace
Readability Assessment & Continuous & 0--100 & Flesch Reading Ease \\
\addlinespace
Politeness Evaluation & Multi-Class & Polite, somewhat polite, neutral, impolite & Bert-base \\
\bottomrule
\end{tabularx}
\end{table}

Sentiment was classified using RoBERTa-base fine-tuned on TweetEval (Negative / Neutral / Positive)\cite{hartmann2022emotionenglish}. Emotion classification followed Ekman’s taxonomy via E-DistilRoBERTa trained on the EmoEvent corpus\cite{plaza2020emoevent} and supplementary datasets (seven categories: disgust, joy, sadness, anger, fear, surprise, neutral)\cite{ekman1992argumet}. Readability was measured with the Flesch Reading Ease score (0–100; higher values indicate greater accessibility). Politeness was evaluated with a BERT-base classifier trained on synthetic customer-service interactions (polite / somewhat polite / neutral / impolite). 

Because callers in emergency contexts are often highly distressed, maintaining clarity and emotional appropriateness is critical. Overly emotional responses may escalate anxiety, while unclear or overly technical instructions risk compromising patient safety. Algorithmic evaluation therefore serves as a safeguard to ensure communication quality alongside clinical safety.

\subsubsection*{Statistical Analysis}

Given skewed category prevalences and multiple raters, we used Gwet’s AC1---less sensitive to prevalence/marginal imbalance than $\kappa$---to estimate inter-rater reliability (95\% CI).
We summarized study outcomes with descriptive statistics (frequencies and percentages) for both binary and ordinal ratings. To examine between-rater differences, we used one-way Analysis of Variance (ANOVA) for ordinal ratings, treating them as approximately continuous and checking normality and homoscedasticity; and Pearson’s Chi-squared tests for binary outcomes (Fisher’s exact test when expected counts were $< 5$). Two-tailed tests were used with $\alpha=0.05$; p-values are nominal with no adjustment for multiple comparisons.


\section{Results}\label{sec3}

This section presents the findings from our hybrid evaluation framework. We first report the results of the human evaluation, including inter-rater reliability and the system's performance on Guidance Efficacy and Dispatch Effectiveness. Subsequently, we present the results from the algorithmic assessment of the agent-generated dialogue.

\subsection{Human Evaluation}\label{subsec_human_results}

\subsubsection*{Inter-Rater Reliability}
To validate the consistency of our human evaluation, we first assessed inter-rater reliability on the 20 cases annotated by all four physicians. The analysis yielded a Gwet’s AC1 score greater than 0.70 across all evaluation metrics, indicating substantial agreement among the physicians and confirming the robustness of our assessment framework. 

\subsubsection*{System Performance on Dispatch and Guidance Efficacy}
\begin{figure}[htbp]
\centering
\includegraphics[width=\linewidth]{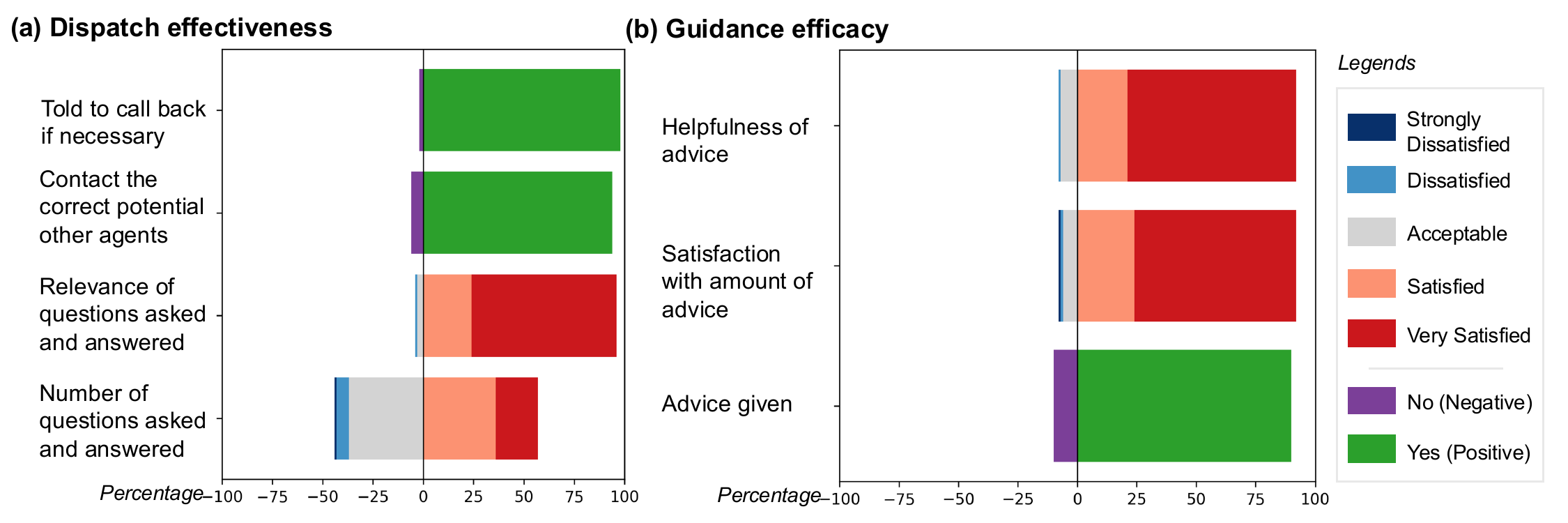}
\caption{Evaluation results in (a) dispatch effectiveness and (b) guidance efficacy. Bars represent the percentage distribution of binary (Yes/No) and ordinal (1–5) responses.}
\label{fig:eval_results}
\end{figure}

The system demonstrated strong performance in both Dispatch Effectiveness and Guidance Efficacy as rated by the physicians. As shown in Figure~\ref{fig:eval_results}(a), the system achieved high Dispatch Effectiveness. In binary assessments, it successfully identified and contacted the correct potential other agents in 94\% of cases and provided advice to call back if necessary in 97\% of cases. For ordinal metrics, the relevance of questions asked was predominantly rated highly (scores of 4 or 5). Notably, the number of questions asked received a neutral rating (score of 3) in 37\% of cases, suggesting an area for further calibration. Guidance Efficacy was rated exceptionally high across all metrics (Figure~\ref{fig:eval_results}(b)). The dispatcher agent provided advice in 91\% of scenarios where it was deemed necessary. Furthermore, both the satisfaction with the amount of advice and the helpfulness of the advice received overwhelmingly positive ratings, with the majority of scores being 4 or 5.

\subsubsection*{Statistical Analysis of Rater Disagreement}

To investigate potential variability among raters, we conducted statistical analyses to compare the physicians' annotations. As summarized in Table~\ref{tab:rater_analysis}, we found statistically significant differences among raters on two metrics: ‘Relevance of Questions Asked and Answered’ (ANOVA, F=4.301, p=0.007) and ‘Contact the correct potential other agents’ (Chi-squared test, $\chi^2$=16.0, p=0.001). For all other metrics, no statistically significant differences were observed, indicating a high degree of consensus. A post-hoc review of individual rater scores revealed that the difference in ‘Relevance of Questions’ was primarily driven by one rater (Physician \#4) providing consistently lower ratings compared to another (Physician \#1).


\begin{table}[htbp]
\centering
\caption{Statistical Analysis of Inter-Rater Variability}
\label{tab:rater_analysis}
\begin{tabularx}{\textwidth}{@{} X l l l @{}}
\toprule
\textbf{Metric} & \textbf{Test Used} & \textbf{Statistic} & \textbf{p-value} \\
\midrule
\multicolumn{4}{l}{\textit{\textbf{Guidance Efficacy}}} \\
Satisfaction with amount of advice & ANOVA & F = 0.458 & 0.713 \\
Helpfulness of advice, if given & ANOVA & F = 1.367 & 0.259 \\
Advice given & Chi-squared & $\chi^2$ = 3.04 & 0.386 \\
\midrule
\multicolumn{4}{l}{\textit{\textbf{Dispatch Effectiveness}}} \\
Number of questions asked & ANOVA & F = 2.264 & 0.088 \\
Relevance of questions asked & ANOVA & F = 4.301 & \textbf{0.007}* \\
Contact correct other agents & Chi-squared & $\chi^2$ = 16.0 & \textbf{0.001}* \\
Told to call back if necessary & Chi-squared & $\chi^2$ = 6.77 & 0.342 \\
\bottomrule
\multicolumn{4}{l}{\small{*Statistically significant difference (p $\le$ 0.05).}}
\end{tabularx}
\end{table}

\subsection{Algorithmic Evaluation}\label{subsec_algo_results}

\begin{figure}[htbp]
\centering
\includegraphics[width=\linewidth]{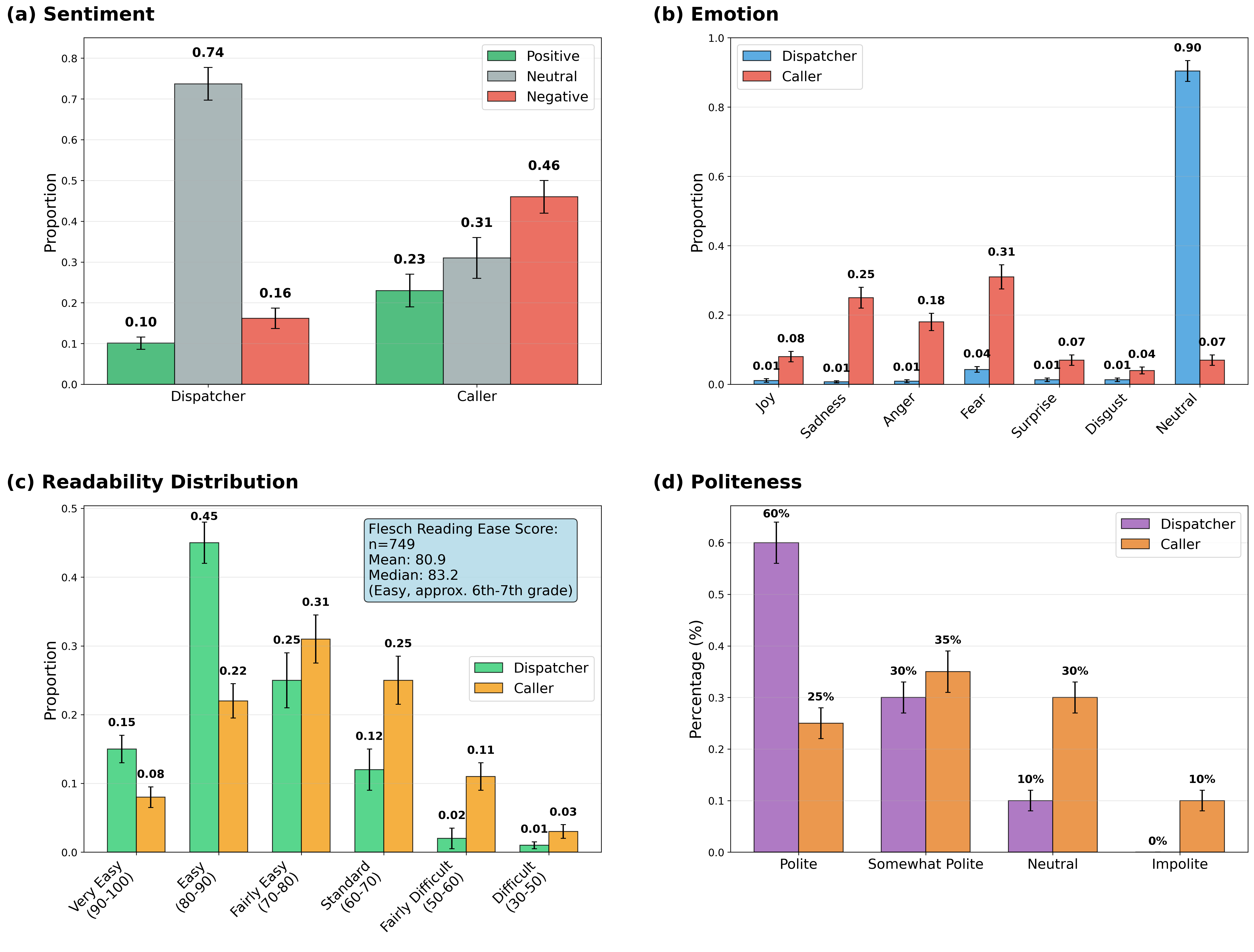}
\caption{Algorithmic evaluation comparing dispatcher and caller dialogue profiles across four key metrics: (a) sentiment distribution, (b) emotion classification, (c) readability based on Flesch Reading Ease scores, and (d) politeness levels.}
\label{fig:algorithmic_results}
\end{figure}

The algorithmic evaluation provided quantitative insights into the communication quality and affective dynamics of the dispatcher-caller dialogues. The analysis revealed a clear and appropriate distinction between the dispatcher's professional, calm profile and the caller's simulated distress, as visualized in Figure~\ref{fig:algorithmic_results}.

\subsubsection*{Sentiment and Emotion Analysis}
Analysis of the agent responses revealed a stark contrast in their affective profiles. As shown in Figure~\ref{fig:algorithmic_results}a, the dispatcher agent maintained a predominantly neutral sentiment (74\%), appropriate for professional emergency communication. In contrast, the caller agent's dialogue was primarily negative (46\%) and less neutral (31\%), effectively simulating the emotional state of a person in an emergency. This emotional gap was further confirmed by emotion classification (Figure~\ref{fig:algorithmic_results}b), where the dispatcher's output was overwhelmingly neutral (90.4\%). Conversely, the caller exhibited significant levels of fear (31\%) and sadness (25\%), emotions consistent with the urgency and seriousness of the scenarios.

\subsubsection*{Readability and Politeness}
The readability of the dispatcher's messages was high, ensuring instructions were clear and accessible to a distressed caller. The messages achieved a mean Flesch Reading Ease score of 80.9, equivalent to a 6th-7th grade reading level. As shown in Figure~\ref{fig:algorithmic_results}c, the distribution of the dispatcher's language peaks in the "Easy" category (43\%), while the caller's language was naturally more varied and complex. Politeness evaluation confirmed a high standard of professional communication from the dispatcher (Figure~\ref{fig:algorithmic_results}d). A majority of its responses were classified as "polite" (60\%), with an additional 30\% as "somewhat polite," and importantly, no responses were deemed "impolite". This demonstrates a consistent adherence to respectful and supportive communication norms, which is critical in high-stakes settings.

\subsection{Operational Performance Dynamics}
To assess the system's efficiency and responsiveness under simulated time pressure, we conducted a post-hoc analysis based on the dialogue transcripts. A quantitative assessment of operational performance across 100 cases, including distributions for information completeness, response timeliness, and guidance accuracy, is presented in Supplementary Figure. 2. The detailed scoring methodology for these metrics is defined in Supplementary Table 10. It is important to note that the timeline used for this analysis is a simulation calculated from the number of utterances, not a measure of real-world computational latency. The core metric, the "Information Collection Efficiency Score," is calculated based on the proportion of predefined critical entities (e.g., location, consciousness) successfully elicited by the agent for a given scenario.

\begin{figure}[htbp]
\centering
\includegraphics[width=\linewidth]{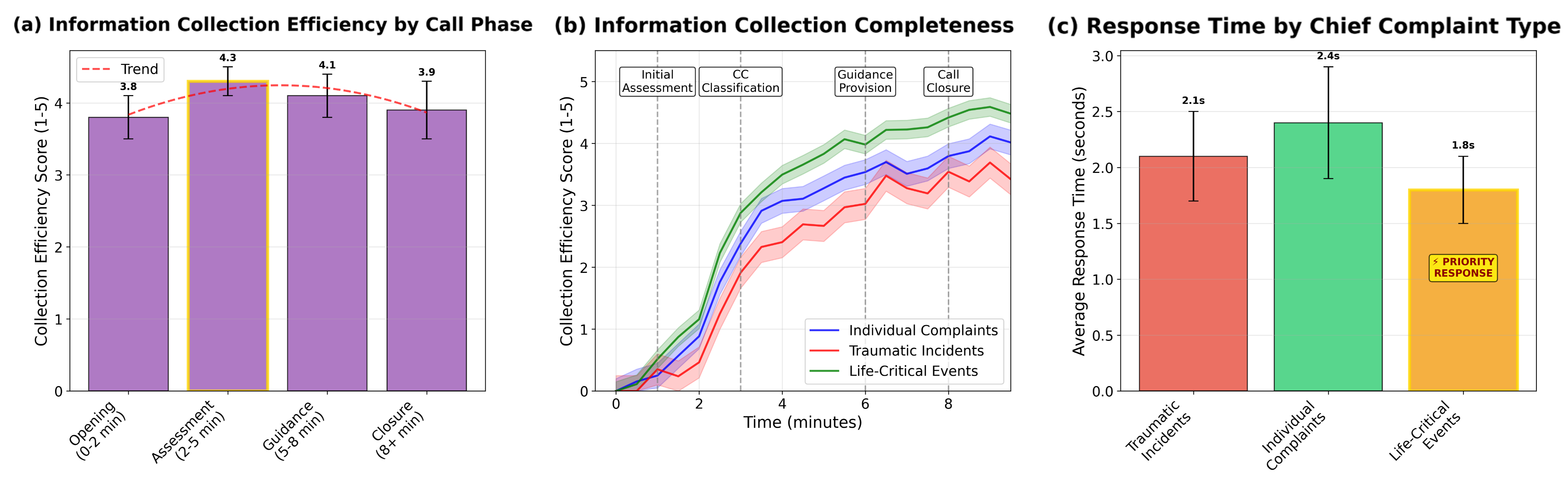}
\caption{Post-hoc analysis of information collection and response dynamics. (a) The Information Collection Efficiency Score (1-5) is highest during the "Assessment" phase of the call, aligning with the core task of that period. (b) The trace of collection completeness over simulated time shows that the system gathers information most rapidly for "Life-Critical Events" (green line), demonstrating effective prioritization. (c) The average simulated response time is fastest for "Life-Critical Events" (1.8s), confirming the system's ability to adapt its conversational strategy to the urgency of the situation.}
\label{fig:stratified_analysis}
\end{figure}

The results, shown in Figure~\ref{fig:stratified_analysis}, indicate that the system’s conversational behavior varies with both the phase of the call and the urgency of the chief complaint. Analysis of information collection efficiency by call phase (Figure~\ref{fig:stratified_analysis}a) reveals an inverted U-shaped trend, with efficiency logically peaking at an average score of 4.3 during the critical "Assessment" phase (simulated 2-5 minutes). Furthermore, when examining performance over the call's duration (Figure~\ref{fig:stratified_analysis}b), the system demonstrated an ability to prioritize urgent cases. The collection completeness for "Life-Critical Events" (green line) increased more rapidly and peaked earlier than for traumatic or individual complaints, indicating an accelerated information-gathering strategy when stakes are highest.

This prioritization was most evident in the system's simulated response time (Figure~\ref{fig:stratified_analysis}c). The agent was significantly faster when handling "Life-Critical Events" (average 1.8s) compared to "Traumatic Incidents" (2.1s) and "Individual Complaints" (2.4s) . This reflects a strategic shift to shorter, more direct, and instructional utterances upon identifying a life-threatening scenario, consistent with the intended prioritization of swift action in the most urgent situations. Analysis of total simulated call duration also confirmed this prioritization, showing a statistically significant difference between groups, with life-critical events having the shortest median duration (Supplementary Figure. 1).

\section{Discussion}\label{sec4}

In this study, we developed and evaluated a novel multi-agent system for simulating emergency medical dispatch, grounded in a comprehensive clinical taxonomy and powered by LLMs. Our hybrid evaluation, combining expert physician assessment with automated linguistic metrics, demonstrated that the system can generate clinically plausible dialogues that adhere to established protocols. The agent-based dispatcher achieved high ratings for both the effectiveness of its operational actions and the efficacy of its pre-arrival guidance. These findings suggest that grounding LLM agents in a structured, domain-specific knowledge framework is a viable and effective strategy for developing safe and reliable AI tools for high-stakes medical environments.

A key finding of this study is the critical role of the taxonomy and fact commons in steering the behavior of the AI agents and mitigating the risk of hallucination. Unlike open-domain chatbots, our agents operate within constraints defined by clinical standards. The system’s ability to dynamically classify CCs and retrieve corresponding protocols at each conversational turn ensured that the dispatcher agent’s responses were not only fluent but also contextually and procedurally appropriate. This approach functionally mirrors the core principle of RAG, where a generative model’s output is anchored to a reliable external knowledge source. By curating a detailed fact commons for each chief complaint, we provided the LLM with a structured ‘source of truth’, substantially improving the safety and predictability of its generated content and establishing a strong foundation for future work incorporating real-time RAG from validated medical databases.

Furthermore, the multi-agent architecture provides significant flexibility for creating realistic and adaptable EMS workflows. This framework moves beyond simple question-answering and simulates the dynamic, multi-party coordination inherent in emergency response. The system's modularity allows for individual components to be independently updated or replaced, such as the caller agent’s personality, the dispatcher agent's reasoning model, or the specific protocols used. This makes it possible to scale the simulation by adding new agents representing other emergency services (e.g., police, fire, poison control) or to integrate external tools and APIs\cite{otal2024llm}, thereby creating a high-fidelity "digital twin" for training, protocol testing, and operational planning.

This inherent flexibility is particularly valuable for addressing the crucial need for social and cultural customization in emergency services. The current framework can be readily adapted to better serve diverse communities. For instance, the LLM's multilingual capabilities can be leveraged to simulate interactions with non-native speakers, helping train dispatchers in cross-cultural communication. The caller identity framework can be expanded to include personas representing various age groups, cognitive abilities, or cultural backgrounds that influence how symptoms are described and how instructions are received. By loading the fact commons with region-specific protocols or public health priorities, the system can be tailored into a highly relevant training tool for different jurisdictions, ultimately promoting more equitable and effective emergency care for all populations. This aligns with broader reviews on the role of LLMs in crisis management and emergency medicine\cite{preiksaitis2024role}.

\subsection*{Limitations and future work}
Despite the promising results, this study has several limitations. First, our evaluation was conducted in a simulated environment, which cannot fully replicate the high-stress, unpredictable nature of real-world 911 calls, including background noise and extreme caller emotional states. Second, while our panel of four physicians provided substantial agreement, a larger and more diverse group of evaluators, including experienced paramedics and professional dispatchers, would provide a more holistic assessment of the system's operational utility. The statistically significant variability observed between two raters on specific metrics highlights the importance of incorporating diverse professional perspectives. Furthermore, to guide iterative refinement, we conducted a qualitative root cause analysis of representative failure cases (detailed in Supplementary Supplementary Method 5), identifying patterns such as 'question overload' and 'premature misclassification' (Supplementary Figure. 4 and Supplementary Table 9). Third, our taxonomy, while comprehensive, is static; a real-world system would need mechanisms to continuously update its knowledge base with the latest clinical guidelines.

Future work will focus on addressing these limitations. Emerging evidence in emergency medicine also highlights the potential of AI-driven approaches\cite{williams2024evaluating}. The next logical step is to conduct prospective validation in a controlled setting with professional dispatchers interacting with the system to assess its utility as a training tool. We also plan to benchmark our system against existing models (e.g., EMSBERT\cite{EMS_BERT}) and on standardized datasets to quantitatively measure its performance. Further development will focus on integrating a true RAG architecture to allow the agent to pull information from live, trusted medical sources. Finally, exploring the complex ethical considerations surrounding AI in emergency medicine, particularly regarding safety, accountability, and the mitigation of algorithmic bias, will be a central theme of our ongoing research.

\section{Conclusion}\label{sec5}

In conclusion, this study demonstrates that combining a structured clinical taxonomy with a multi-agent LLM framework enables the simulation of complex EMS dispatch scenarios in a clinically grounded and operationally realistic manner. The proposed approach provides a blueprint for developing, validating, and customizing AI agents to support high-stakes medical communication. Beyond serving as a platform for next-generation training and decision support tools, this work highlights a pathway toward the safe and effective integration of AI into emergency response workflows. While further validation in real-world dispatch environments remains essential, our findings establish a foundation for advancing AI-assisted emergency medicine in both research and practice.

\section*{CRediT authorship contribution statement}

\textbf{Xiang Li}: Conceptualization, Data curation, Formal analysis, Investigation, Methodology, Visualization, Writing\,\textendash{} original draft, Writing\,\textendash{} review \& editing.
\textbf{Huizi Yu}: Conceptualization, Data curation, Formal analysis, Investigation, Methodology, Visualization, Writing\,\textendash{} original draft, Writing\,\textendash{} review \& editing.
\textbf{Wenkong Wang}: Conceptualization, Data curation, Formal analysis, Investigation, Methodology, Visualization, Writing\,\textendash{} original draft, Writing\,\textendash{} review \& editing.
\textbf{Yiran Wu}: Data curation, Formal analysis, Investigation, Methodology.
\textbf{Jiayan Zhou}: Data curation, Formal analysis, Investigation, Methodology.
\textbf{Wenyue Hua}: Data curation, Investigation, Writing\,\textendash{} review \& editing.
\textbf{Xinxin Lin}: Data curation, Investigation, Writing\,\textendash{} review \& editing.
\textbf{Wenjia Tan}: Data curation, Investigation, Writing\,\textendash{} review \& editing.
\textbf{Lexuan Zhu}: Data curation, Investigation, Writing\,\textendash{} review \& editing.
\textbf{Bingyi Chen}: Data curation, Investigation, Writing\,\textendash{} review \& editing.
\textbf{Guang Chen}: Data curation, Investigation, Methodology.
\textbf{Ming-Li Chen}: Data curation, Investigation, Methodology.
\textbf{Yang Zhou}: Data curation, Investigation, Methodology.
\textbf{Zhao Li}: Data curation, Investigation, Methodology.
\textbf{Themistocles L.\ Assimes}: Conceptualization, Methodology, Writing\,\textendash{} review \& editing.
\textbf{Yongfeng Zhang}: Conceptualization, Methodology, Writing\,\textendash{} review \& editing.
\textbf{Qingyun Wu}: Conceptualization, Methodology, Writing\,\textendash{} review \& editing.
\textbf{Xin Ma}: Conceptualization, Investigation, Supervision, Writing\,\textendash{} review \& editing.
\textbf{Lingyao Li}: Conceptualization, Investigation, Supervision, Writing\,\textendash{} review \& editing.
\textbf{Lizhou Fan}: Funding acquisition, Conceptualization, Data curation, Formal analysis, Investigation, Methodology, Supervision, Writing\,\textendash{} original draft, Writing\,\textendash{} review \& editing.

\section*{Code availability}
The complete code is available at
\url{https://doi.org/10.5281/zenodo.17420907}. 

\section*{Acknowledgements}

The authors acknowledge the following funding sources: the Chinese University of Hong Kong Vice-Chancellor Early Career Professorship Scheme (L.F.), National Natural Science Foundation of China (Grant No. 62573271) (X.M.), and Major Basic Research Project of Shandong Provincial Natural Science Foundation (Grant No. ZR2025ZD28) (X.M. and L.F.). The content is solely the responsibility of the authors and does not necessarily represent the official views of universities or hospital authorities. 

\section*{Declarations}

\subsection*{Ethics approval and consent to participate}
This study was approved by Shandong University Qilu Hospital Ethics Committee Review Board (IRB No. KYLL-202505-005) and conducted in accordance with the principles of the Declaration of Helsinki. The requirement for informed consent was waived due to the retrospective analysis of publicly available anonymized data.

\subsection*{Competing interests}
The authors declare that they have no known competing financial interests or personal relationships that could have appeared to influence the work reported in this paper.

\appendix
\section*{Appendix A. Supplementary data}

Supplementary material related to this article can be found online at 
\href{https://doi.org/10.5281/zenodo.17420907}{https://doi.org/10.5281/zenodo.17420907}.

\bibliographystyle{elsarticle-num} 
\bibliography{references}

\end{document}